\pdfoutput=1
\documentclass[11pt]{article}

\usepackage[]{acl}

\usepackage{times}
\usepackage{latexsym}

\usepackage[T1]{fontenc}
\usepackage[utf8]{inputenc}

\usepackage{microtype}

\usepackage{inconsolata}

\usepackage{array}
\usepackage{booktabs}
\usepackage{multirow}
\usepackage{amsmath}
\usepackage{amssymb}
\usepackage{graphicx}
\usepackage{enumerate}
\usepackage{diagbox}
\usepackage{enumitem}
\usepackage{color}
\usepackage{xcolor}
\usepackage{soul}

\newcommand{\tabincell}[2]{\begin{tabular}{@{}#1@{}}#2\end{tabular}}

\title{What Really is Commonsense Knowledge?}
\author{
\textbf{Quyet V. Do, Junze Li, Tung-Duong Vuong, Zhaowei Wang,}\\
\textbf{Yangqiu Song, Xiaojuan Ma}\\
Department of Computer Science and Engineering, HKUST, Hong Kong SAR, China \\
\texttt{\{vqdo, junze.li, tdvuong, zwanggy\}@connect.ust.hk},\\
\texttt{\{yqsong, mxj\}@cse.ust.hk}
}

\begin{document}
\maketitle

\begin{abstract}
Commonsense datasets have been well developed in Natural Language Processing, mainly through crowdsource human annotation.
However, there are debates on the genuineness of commonsense reasoning benchmarks. In specific, a significant portion of instances in some commonsense benchmarks do not concern commonsense knowledge.
That problem would undermine the measurement of the true commonsense reasoning ability of evaluated models.
~\citet{davis2023survey} suggested that the problem originated from a blurry concept of commonsense knowledge, as distinguished from other types of knowledge. 
To demystify all of the above claims, in this study, we survey existing definitions of commonsense knowledge, ground into the three frameworks for defining concepts ~\cite{murphy2004concept}, and consolidate them into a multi-framework unified definition of commonsense knowledge (so-called consolidated definition).
We then use the consolidated definition for annotations and experiments on the CommonsenseQA and CommonsenseQA 2.0 datasets to examine the above claims.
Our study shows that
% \Red{(tentative assumption) MTurk annotators or crowdsource annotators in general have perspectives of the concept commonsense different from experts}, 
there exists a large portion of non-commonsense-knowledge instances in the two datasets, and a large performance gap on these two subsets where Large Language Models (LLMs) perform worse on commonsense-knowledge instances.
\end{abstract}

% Working definition of Commonsensen Knowledge: 1) the inductive knowledge learned through experience and observation as such knowledge are unlikely to be published in textbooks, dictionaries, magazines, or encyclopedias, even those designed for children (Lenat 1995, Liu and Singh 2004), 2) commonly believed but not guaranteed to be true (Davis and Marcus, 2015).

\section{Introduction}

Commonsense datasets have been well developed in Natural Language Processing since the last decade. As commonsense data is known to be implicit, almost all commonsense datasets are constructed through crowd source human annotation instead of relying on automated dataset construction processes.
These commonsense datasets serves as valuable resources to augment AI models in various aspects, such as text generation~\cite{zhou2021commonsense, ilievski2021story}, visual reasoning ~\cite{zellers2019vcr},
or building more capable knowledge models for further downstream applications ~\cite{yu2022cocolm, hwang2021comet, wang-etal-2023-cat}, as well as benchmarks to evaluate the reasoning capability of AI models ~\cite{talmor-etal-2019-commonsenseqa, zhang-etal-2020-winowhy, bhagavatula2020abductive, talmor2022commonsenseqa, fang2023ckbpv2}.

However, there are debates on the quality of commonsense datasets, especially when they serves as evaluation benchmarks.
~\citet{davis2023survey} argued that many prevalent commonsense datasets are flawed in the sense that they contained a significant portion of instances which do not concern \textbf{commonsense} knowledge but other types of knowledge, namely common, encyclopedia, and expert knowledge (so-called \textbf{referenced} in this work).
For example, in the CommonsenseQA 2.0 dataset ~\cite{talmor2022commonsenseqa} which consists of Yes/No questions (or assertions), the instance ``A male seahorse cannot give birth'' (Answer: no) presents common biology knowledge, meanwhile, ``Electrons are smaller than mesons'' (Answer: no) is certainly an encyclopedic fact.
As it has been widely discussed that language models excel in memory or retrieval tasks while still struggle in reasoning tasks ~\cite{bang2023chatgpt, yoav2023llmremark, huang-chang-2023-towards},
% It suggests that tasks involving commonsense knowledge (or in a possibly overclaimed term, commonsense reasoning tasks) are more challenging than tasks involving referenced knowledge, which concerns memory retrieval more.
% https://community.openai.com/t/foundational-must-read-gpt-llm-papers/197003/14
the flaw in commonsense datasets would undermine the measurement of the true commonsense reasoning ability of evaluated models.
% i.e. the ability of the ``core language understanding and reasoning'' component.
~\citet{davis2023survey} suggested that the problem originated from a blurry concept of commonsense knowledge, as distinguished from other types of knowledge. Due to that blurry concept, both annotators and researchers working on {commonsense} may not be aware of the genuineness problem of commonsense datasets.

% brief literature review
Indeed, according to our literature review, all works on {commonsense} have their ways to describe the concept. However, the description of each is not adequate and comprehensive to educate outsiders or even insiders of this research field about the concept and the difference with respect to other relevant concepts, such as {referenced} knowledge.
% Also, through the lens of three major frameworks to define a concept (namely \textit{ideal, feature,} and \textit{exemplar}) which are discussed in The Big Book of Concept~\cite{murphy2004concept}, {we posit that the existing literature on {commonsense} only provides a limited number of features which are even scattered among different works, making the concept {commonsense} not systematically and comprehensively depicted.}
Also, through the lens of concept definition theory ~\cite{murphy2004concept}, {we posit that each of the existing literature on {commonsense} only provides a limited number of features, making the concept {commonsense} not systematically and comprehensively depicted.}
% Regarding the problems,~\cite{murphy2004concept} suggested 1) the importance of one item's typicality w.r.t. both the target and alternative concepts in object categorization, thus the importance of the separation of one concept from other concepts,
% Also, each of these Views is not self-sufficient and has its own weakness to represent a concept (none of the existing views suffices to represent a concept).
% and 2) in different cases, one definition framework will have to be combined with one of the other frameworks in order to form a complete theory of concepts, e.g. exemplar framework side-by-side with feature framework in the process of learning a new concept.
Regarding the problems,~\citet{murphy2004concept} suggested combining general description, examples, and list of features of the concept to form a complete definition.

% Therefore, the consolidation of existing literature in the two aforementioned aspects are important to ``educate'' researchers and annotators about {commonsense}, which ultimately helps to produce more genuine commonsense benchmarks to measure the true commonsense reasoning capability.

Motivated by that research gap, in this work, we consolidate the definition of {commonsense} as follow. Firstly, leveraging the descriptions about commonsense and referenced knowledge from previous works, we attempt to distinguish commonsense knowledge from referenced knowledge. We provide a table of representative cases for each concept to \textit{show} (or more subjectively, \textit{assume}) the fundamental and subtle difference of these two concepts.
% by systematically grounding definition of commonsense in previous works into the major views.
Based on the descriptions and examples,
% \footnote{That is to say people may have their own definitions of commonsense which are different from the unified one. Therefore, our proposed prototype view, as conditioned on the unified one, may not fit everyone's perspective.},
we {systematically} propose a list of multi-aspect binary-value features that characterize commonsense, referenced knowledge, and their difference. We then validate the significance of features through an empirical study on the CommonsenseQA~\cite{talmor-etal-2019-commonsenseqa} and CommonsenseQA 2.0~\cite{talmor2022commonsenseqa} datasets. Overall, we observe that 1) whether we can obtain the knowledge by our own experience/observation and 2) whether the knowledge is only mutual belief are the most significant features to identify commonsense from referenced knowledge.

Given the consolidated definition, we 
% study the perspective of MTurk annotators toward commonsense knowledge as distinguished from referenced knowledge, comparing to expert's perspective which reflects our consolidated definition, through their annotation on the CommonsenseQA and CommonsenseQA 2.0 datasets.
% with two control settings - simple description of commonsense from each dataset and the consolidated definition.
% The result show that \Red{(tentative assumption) MTurk annotators or crowdsourced annotators in general have perspectives of the concept "commonsense" different from experts, which likely lead to the aforementioned genuineness problem of commonsense datasets}.
% , and having a consolidated definition of commonsense knowledge helps to align their perspectives to that of expert.} % (the most reasonable perspective) which might be a bit trivial.
% Furthermore, we apply the consolidated definition to 
analyze the portion of instances of commonsense and referenced knowledge in the development sets of the CommonsenseQA and CommonsenseQA 2.0 datasets, as well as the performance of Large Language Models (LLMs) such as Gemini-Pro, ChatGPT, LLaMa2-7B, and Mixtral-8x7B on the commonsense-knowledge subsets (which consists of instances required commonsense knowledge to answer) and referenced-knowledge subsets from these datasets. Aligned with the claims which motivate for this work, we observe a large portion of referenced knowledge in the two datasets (0.27 ± 0.09 for CommonsenseQA and 0.56 ± 0.1 for CommonsenseQA 2.0)\footnote{95\% confidence interval}, and a large performance gap (4 to 7 point of accuracy) on these two subsets, where LLMs perform worse on commonsense-knowledge instances, suggesting that commonsense reasoning tasks or reasoning tasks in general are more challenging than memory-retrieval tasks.

% Overall, our contributions are as follow: ...
The organization of this paper is as follows. In section 2, we discuss related works (especially background knowledge on the frameworks to define a concept in ~\citet{murphy2004concept}). In section 3, we show a comprehensive survey on the definitions of commonsense knowledge
% across foundation, prevalent, and relevant previous works 
by grounding relevant previous works into three aforementioned definition frameworks, then provide a table of representative cases. After that, we describe the list of features and validation procedure.
Finally, in section 4, we apply the consolidated definition to demystify relevant claims which motivate this work.

\section{Background and Related Works}

\paragraph{Commonsense}
% Don't further discuss about definition here 

Research on commonsense had been started from the last century and early 2000s, with the foundational works such as~\citet{davis1990-iv, Lenat1995-vm, jmc2002commonsense, Liu2004-ag}.
They laid the first building blocks on the definition, meaning, and practical application of commonsense knowledge in the field of language processing. % four works respectively answered/stated "involvement of commonsense knowledge and reasoning in intelligent activities", "surveys the range of applications (of such a large-scale project)", "what is commonsense", and "large dataset"
Recent time has witnessed a drastic development of research in commonsense knowledge and reasoning, including resources and benchmarks, with the wide range of format (free text, knowledge graph, knowledge bases, etc.), topics (concept taxonomy, geographical tradition, daily inference, etc.), evaluation tasks (abductive reasoning, question answering, or generation), as well as semantics dimensions (physical, linguistic, textual worlds, etc.)~\cite{ilievski2021dimensions}.
Despite a number of works on commonsense, in section 3, we show that the definition of commonsense in existing works is not comprehensive, which possibly leads to the quality problem of commonsense datasets.

% in the past
% recently: topic, task and evaluation (need more citation, check Mendeley, ask people)
% yet in Section 3, we posit that the definition of commonsense in existing work is blurry.

\paragraph{Definition Frameworks}
% Concept ~ Mental representation
% Contegory ~ Class of object in the world
% Talk about 4 types of view, and 3 core frameworks

As introduced in the \textit{The Big Book of Concepts}~\cite{murphy2004concept}, there are three major frameworks of how we understand and organize the world around us through the notion of concepts. These frameworks motivate four main Views of concept: Classical View, Prototype View, Exemplar View, and Knowledge View. In layman term, a View of concept means a theoretical way to define a concept.
% how human formulate a class of objects and use it to recognize objects of that class.
% Classical View~\cite{hull1920quantitative, smoke1932concept}, one of the earliest and fundamental views of categories, yet its strictness in member categorization that one item can either belong to a category or not leads to some unsolved cases such as penguin is a bird, but bird is defined as a species that can fly. Therefore, three other views were introduced to complement the Classical View.

Motivating the Knowledge View~\cite{murphy1985coherence}, the \textit{ideal} framework~\cite{barsalou1985ideals} views each concept as a part of our knowledge and understanding of the world, in which we do not learn concepts in isolation. It describes how each concept fits in other parts of our lives, e.g. what is its meaning, how to use it, why people create it, etc. For examples, through the \textit{ideal} framework, weapon is defined as a thing ``designed or used for inflicting bodily harm or physical damage''.
The \textit{feature} framework, which bases Classical and Prototype Views, represents a concept by a list of its most typical features. In item categorization, one may examine the similarity of the item to the feature list, then for every feature the item has, it gets ``credit'' for the feature's weight, accumulating to the typicality of the item w.r.t. to the concept. Features are in fact not necessarily disjoint in term of semantics. For example, weapon can be characterized by three features ``can do harm'', ``made of metal'', and ``sharp''.
In a more complex setting (e.g. schemata), value of each feature can be nominal or continuous. However, in this work, we only consider a simple setting with binary-value feature. % Each feature is weighted according to its importance in representing the concept. That's what we will do in linear regression model
In contrast, the \textit{exemplar} framework, as the core of Exemplar View, represents a concept through a collection of specific instances or exemplars that exemplify the concept. This framework emphasizes the role of individual examples in categorization and facilitates greater flexibility in category boundaries.
Overall, each framework concerns a level of abstraction of a concept, and one framework is likely not adequate to represent the concept.

% borrow words from Conclusion (page 64, or 73/564)
% (copy from page 74/..) Thus, exemplar knowledge and prototype knowledge must exist side by side to at least some degree, according to prototype theory. The general claim of that theory, however, is that for mature categories, people rely on summary representations of the entire category rather than specific exemplars in making judgments about the concept.

\paragraph{Analysis with Categorization} %(see more and more papers). Data categorization for Analysis is what occurs in this work, but not really in other papers. In other papers, analysis is the main contribution. This work, since there is no clear categorization yet, thus contribute in that aspect to facilitate later analysis
By nature, the human knowledge is expanded through the separation of concepts for deeper study.
Many developments of different fields, e.g. AI~\cite{huang-chang-2023-towards, sun2023headtotail, wang2023far}, cognitive science~\cite{genon2018characterize}, medicine~\cite{stein2013psychiatry}, etc., follow the pattern to offer deeper understanding of existing problems and better solutions.
Recently, in the field of of AI, there are more and more efforts put into such type of analytical works.
For example, to study the design bias of datasets and models in term of their creators' identity and background,~\citet{santy-etal-2023-nlpositionality} introduced a framework to quantify the alignments of human subjects of different demographic categories with datasets' labels and model predictions.
% Meanwhile,~\cite{sun2023headtotail} examines how LLMs are knowledgeable regarding to entities of different popularity in the reference corpora.
% In another aspect,~\cite{ye2023flask} introduces a pool of 12 fine-grained skills needed for LLMs to follow open-ended user instructions and comprehensively analyzes LLMs' capability with respect to instance-level required skills, domains, and difficulty. 
Likewise,~\citet{fang2023getting} systematically define distinct kinds of bias in event temporal reasoning, then study knowledge conflicts arising from mismatches between actual temporal relations of events and the prior biases learned by the model.
Similarly, in this work, we define commonsense as distinguished from % (the "correct" expression I learned from the big book of concept)
referenced knowledge to better understand how LLMs perform in instances regarding each type of knowledge.
% In contrast to categorization of aforementioned works, the categorization of commonsense as distinguished from referenced knowledge seems rather more subjective, unclear, and unfortunately not widely discussed in previous works on commonsense. Therefore, this work posits its main contribution on the categorization of commonsense and commonsense knowledge, and minor contribution on the analysis of LLMs' performance w.r.t the two types.

\paragraph{Terminology}
To avoid ambiguity, we explain some terminology used in this work. The word ``commonsense'' refers to commonsense knowledge (e.g. Mountains are high), and ``non-commonsense'' refers to knowledge of another type (e.g. Washington D.C. is the capital of the US) rather than an incorrect commonsense knowledge (e.g. PersonX likely goes to sleep when he is hungry).
Furthermore, ``knowledge type of an instance'' refers to the type of the knowledge used to do the corresponding task with the instance.

% Furthermore, because of the possible difference and independence of knowledge and inference~\cite{zhang-etal-2023-cikqa, davis2023survey}, we clarify that we work on the definition of commonsense knowledge but not that of commonsense reasoning.
% That means a task instance which concerns referenced knowledge by our consolidated definition may be still claimed to test commonsense reasoning ability by others. % researchers.

\section{Commonsense vs. Referenced Knowledge}
%..., in comparison to ...

\subsection{``Commonsense'' in Previous Works}
\label{sec:review}

\begin{table*}[h!]
\small
\centering
\setlength{\tabcolsep}{4.5pt} % Default value: 6pt
\renewcommand{\arraystretch}{1} % Default value: 1
\resizebox{\linewidth}{!}{%
\begin{tabular}{p{0.1\linewidth} | p{0.1\linewidth} | p{0.35\linewidth} | p{0.25\linewidth} | p{0.2\linewidth}}

\toprule
Prev. Work & Topic & \textit{Ideal} & \textit{Feature} & \textit{Exemplar}
\\

\midrule % foundation (Cyc, WSC, ConceptNet)
% \midrule
% The Structure of a Semantic Theory &  & MIT & % very hard to read!
% &
% &
% &
% \\
\midrule
\citet{Lenat1995-vm} (CYC) & General common sense resource
& "... Stating them [commonsense assertions] to another person, aloud or in print, would likely be confusing or insulting ..."
& "Such assertions are unlikely to be published in textbooks, dictionaries, magazines, or encyclopedias, even those designed for children."
& "You have to be awake to eat", "You can usually see people’s noses, but not their hearts", etc. %\tabincell{l}{}
\\

\midrule
\citet{jmc2002commonsense} & General description of common sense ability 
& ".. Common sense involves certain abilities to decide what to do to achieve goals .. We may call the human ability to take facts into account common sense reasoning ability .."
& ".. The facts that describe the consequences of events the actor doesn’t control, are the most important common sense knowledge .."
& "London is in the south of England.",
"When objects collide they usually make a noise"
\\

\midrule
\citet{Davis2015-so} & Theory
& "Piecemeal commonsense knowledge (e.g. specific facts) is relatively easy to acquire, but often of little use, because of the long-tail phenomenon discussed above"
& "Commonsense reasoning almost always involves plausible reasoning; that is, coming to conclusions that are reasonable given what is known, but not guaranteed to be correct"
& "If you see a six-foot tall person holding a two-foot tall person in his arms, and you are told
that they are father and son, you do not have to ask which is which"
\\

\midrule
\citet{davis2023survey} & Survey on common sense resources and benchmarks
& "Common sense supports reasoning", "Commonsense knowledge can be distinguished from from referenced knowledge, encyclopedic knowledge and expert knowledge"
%, "Commonsense reasoning is integrated with other cognitive abilities",
% "common sense knowledge is independent of any specific task or modality."
& "Common sense is largely sensible", "Commonsense knowledge is not book learning, explicitly taught in schools."
& "The sun is very bright", ".. People cannot walk from one [Central Park] to the other [the Golden Gate Bridge] in fifteen minutes .."
\\

\midrule
\citet{Liu2004-ag} (ConceptNet) & General commonsense resource
& %"While to the average person the term `commonsense' is regarded as synonymous with `good judgement', "
"To the AI community it [commonsense] is used in a technical sense to refer to the millions of basic facts and understandings possessed by most people"
& ".. Such [commonsense] knowledge is typically omitted from social communications, such as text."
& "A lemon is sour", "To open a door, you must usually first turn the doorknob", %"If you forget someone’s birthday, they may be unhappy with you"
\\
% Reporting Bias and Knowledge Acquisition & JHU &  & % seem not good ref, as only focus on reporting bias (no discrimination or specific description of commonsense vs referenced knowledge)
% & 
% & commonsense isn’t usually stated explicitly 
% & "People use money to buy things"
% \\

\midrule % emergence (UW/AllenAI ignite a new branch of research with ATOMIC)
\midrule
\citet{sap2019atomic} (ATOMIC) & Daily inferential knowledge 
& "[Commonsense reasoning is] about what might have happened just before, what might happen next as a result, and how different events are chained through causes and effects"
& n/a
& "X repels Y’s attack because X wanted to protect herself", "PersonX makes PersonY’s coffee thus PersonX [want to] adds cream and sugar"
\\

\midrule
\citet{talmor-etal-2019-commonsenseqa} (CommonsenseQA) &  Common sense resources and benchmarks
& "When humans answer questions, they capitalize on their common sense and background knowledge about spatial relations, causes and effects, scientific facts and social conventions."
& n/a
& "When Simon heard the lawn mower, he was probably  outdoors and situated at street level"
\\

\midrule
\citet{sap-etal-2019-social} (SocialIQA) & Social and emotional intelligence 
& "Social and emotional intelligence [as commonsense knowledge] enables humans to reason about the mental states of others and their likely actions"
& n/a
& "Alex spilled the food she just prepared all over
the floor and it made a huge mess. Alex will want to mod up."
% , "In the school play, Robin played a hero in the struggle to the death with the angry villain. Others will be hopeful that Robin will succeed", 
\\

\midrule % derivative (other universities)
\midrule
\citet{onoe2021creak} (CREAK) & Entity understanding and common sense inference.
& "These concepts [commonsense about everyday scenarios (physical, social, etc.) and factual knowledge about entities] overlap in a set of inferences involving entities that we call entity commonsense."
& n/a
& "If you’re good at a skill you can teach others how to do it"
\\

% \midrule
% Commonsense Reasoning for Conversational AI: A Survey of the State of the Art & Conversational AI 
% & Commonsense knowledge is generally understood as external knowledge about the world that all humans are assumed to possess (Liu and Singh, 2004).
% & n/a
% & n/a
% \\

\midrule % Hongming also wants to make clear of knowledge and inference/reasoning capability!
\citet{zhang-etal-2023-cikqa} (CIKQA) & Common sense benchmarks
& "Understanding human language requires both the language knowledge (e.g., grammar and semantics) and world knowledge, which can be further divided into factual and commonsense knowledge (Katz and Fodor, 1963). "
& n/a
& "I drank from the water fountain. The cause of this was I was thirsty.", "The fish ate the worm. It was hungry. That means the fish was hungry."
\\

\bottomrule
\end{tabular}
}
\caption{Literature review on previous works on commonsense. We directly quote descriptions about commonsense for credibility. Text in [] means to complete the quotes. N/a means ``not exist or copied/adapted from previous work''. For works concerning QA tasks, we convert instances from the QA format to free-text format.}
\label{tab:literature}
\vspace{-0.1in}
\end{table*}

% terminology: definition framework, including defining ideal, defining feature, defining exemplar
% three definition frameworks, under three groups of works
A summary of literature review on many previous works on commonsense in term of three definition frameworks - \textit{ideal, feature}, and \textit{exemplar} - is shown in Table \ref{tab:literature}. In the table, we show representative works from three groups: 
1) decade-old-foundational or theoretical works, % However, there is so far no paper attempting a general description of what common sense ability is. (McCarthy, 2002) -> works on commonsense before 2000 did not well describe the term
2) works from the Allen Institute for Artificial Intelligence (AI2)\footnote{Allen Institute for Artificial Intelligence has made commonsense reasoning a major focus of many of its research projects, as sponsored by DARPA \cite{davis2023survey}}, 
3) works from a variety of research groups.

% phenomenon about definition and citation flow
In general, except decade-old-foundational or theoretical works, all other works that we surveyed ~\cite{sap2019atomic, talmor-etal-2019-commonsenseqa, sap-etal-2019-social,  zellers-etal-2018-swag, zellers2019hellaswag, hwang2021comet, Lourie2021UNICORNOR, forbes-etal-2020-social, onoe2021creak, zhang-etal-2023-cikqa, richardson2023commonsense, madaan-etal-2022-language, sun-etal-2022-improving, maharana-bansal-2022-curriculum, lu2023neurosymbolic, yin-etal-2022-geomlama, porada-etal-2022-pre, zhou-etal-2022-think, qasemi-etal-2022-paco, DBLP:conf/aaai/ZhouZCH20} do not provide any defining feature about the concept ``commonsense''.
They rather provide minimum description (varying from one paragraph to one sentence) as domain-specific ideal about commonsense. 
That is a common phenomenon in reporting, as humans tend not to fully express a term defined before. Indeed, aforementioned works rely on previous works for reference about the concept. % in a broader sense.
Interestingly, we observed that almost all works in group 2 and 3 refer the concept ``commonsense'' back to \citet{Liu2004-ag}, and a majority of works in group 3 refer it back to or cite works in group 2.
In term of five mentioned decade-old-foundational or theoretical works, each paper did define commonsense using all definition frameworks, yet each posed at most one or two features of commonsense knowledge, while the rest of the description about commonsense falls into definition frameworks \textit{ideal} or \textit{exemplar}.
That makes our claim about the lack of a systematic and comprehensive list of defining features of commonsense knowledge.

Furthermore, while \textit{ideals} about commonsense of these works (except \citet{davis2023survey}) represent commonsense indistinguishable from basic facts, even including referenced knowledge~\cite{Halpern1990-ng} such as law, regulation, convention, natural science, etc.;
\textit{feature} and \textit{exemplar} views of these works (except \citet{Liu2004-ag} and \citet{jmc2002commonsense} respectively) seems not to represent such referenced knowledge.
That means \citet{Liu2004-ag} and \citet{jmc2002commonsense} treat commonsense as general basic knowledge which is indistinguishable from referenced knowledge, and three other works \cite{Lenat1995-vm, Davis2015-so, davis2023survey} to some extent treat commonsense different from referenced knowledge.
Given that almost all other surveyed works follow the definition of commonsense in \citet{Liu2004-ag}, we conclude that almost all previous works on commonsense do not consider the difference between commonsense and referenced knowledge.

% AI2, leading organization in the field of commonsense, funded by DARPA, have many works that build commonsense data resources, but the definition or description about the term commonsense is limited, relied on foundational work. Prob. ConceptNet. Their ideal about commonsense are more domain specific, containing no or little semantics of commonsense. Also not to mention they pose no feature of commonsense (as reference for annotator), thus the understanding about commonsense is in fact also a commonsense knowledge. 

At this moment, a researcher would have two directions regarding this problem.
One is to continue to treat commonsense and referenced knowledge similar, as ``the boundaries between these categories are extremely vague and it would be wasted effort to try to make them precise ... [then] we are just studying the use of world knowledge generally, or reasoning generally''~\cite{davis2023survey}. 
The other is to consider the subtle difference of commonsense and referenced knowledge and close this theory gap by de-blurring the boundary between commonsense and referenced knowledge.
In this work, we incline to the latter. According to the reporting bias, we argue that referenced knowledge is inherently reported more frequently (even long-tail referenced knowledge will be reported) and consistently (e.g. laws must be reported with integrity) than commonsense knowledge.

% TODO: do we need to mention citation regarding referenced knowledge anymore? No. Actually there is only one work really makes the point of referenced knowledge or reference knowledge

\subsection{Descriptions and Examples}
% Use slide about summary of cases
% Use the table

We give a summary of assertions that concern either commonsense or referenced knowledge in Table \ref{tab:assumption} (with reference to previous works), as the mutual ground between us and readers to base our proposal of a list of binary-value features that distinguishes commonsense from referenced knowledge. % NOTE: Once again, we note that we only consider assertions (free-text format) as our produced theory is generalizable, while tasks may require some tricks which are not involved in the underlining knowledge.
% We follow the convention that commonsense is implicit, while referenced knowledge is explicit.

\begin{table*}[h!]
\small
\centering
\setlength{\tabcolsep}{4.5pt} % Default value: 6pt
\renewcommand{\arraystretch}{1} % Default value: 1
\resizebox{\linewidth}{!}{%
\begin{tabular}{p{0.12\linewidth} | p{0.44\linewidth} | p{0.44\linewidth}}

\toprule
Aspects $\downarrow$ & \tabincell{c}{Commonsense} & Referenced 
\\

\midrule
Linguistics
& N/A
&
\begin{tabular}{p{.95\linewidth}}
\textbf{Definition or linguistic meaning of a word}\\
Reference: \citet{davis2023survey} \\
Example: "If a person is trying to keep something in their hand, they should hold it".\\
(Elaboration: "hold" means "keep something in hand", hence this is the definition of the word)
\end{tabular}
\\

\midrule
Characteristics of Objects / Entities
&
\begin{tabular}[t]{p{.95\linewidth}}
\textbf{Non-defining characteristics of general objects}\\
Reference: \citet{jmc2002commonsense} \\ % commonsense knowledge fill up the knowledge space 
Example: "Mountains are high".\\
(Elaboration: This is not necessary that a general mountain is high, but it is sensibly true. Yet the highness of a mountain is dependent of observers)
\end{tabular}
&
\begin{tabular}{p{.95\linewidth}}
\textbf{Characteristics of a specific entity}\\
Reference: \citet{Halpern1990-ng} \\
Example: "The Fuji Mountain is 3,776 metres high".\\
(Elaboration: The Fuji Mountain is a specific entity, thus the statement is verifiable)\\
\\
\textbf{Defining characteristics of general objects}\\
Reference: Derived from the word "referenced" \\
Example: "A mountain is a large natural elevation of the earth's surface rising abruptly from the surrounding level".\\
(Elaboration: This is the definition of "mountain" from Google Translate)
\end{tabular}
\\

\midrule
Facts, Actions, States, and Events
&
\begin{tabular}{p{.95\linewidth}}
\textbf{Implicit mutual belief (which is rarely conventionalized)}\\
Reference: \citet{Liu2004-ag, davis2023survey} \\ % knowledge generally possessed but sensible
Example: "At a funeral, a person would be sad".\\
(Elaboration: It is a social etiquette to empathize with the lost of the family, but there is no written-down convention of the etiquette) \\
% Example: "Teachers must deal with a lot of paperwork".\\
% (Elaboration: As teachers are obligated to deal with class preparation, exam preparation, grading, etc. which involve in paperwork, we mutually believe teachers need to deal a lot of paper work. However, it is not conventionalized as a duty of teacher.) \\ % However, some nearly retired teachers may not need to do so, and the phrase 'a lot of work' is subjective
\\
\textbf{Derivative of facts which is not written down and not always true.}\\
Reference: \citet{Davis2015-so} \\
% Example: "The capital of a country should be a region with a large number of residents".\\
% (Elaboration: There is no such constitution to set the capital. Also, for some countries like Australia or Canada, the capital is far less famous than other cities, and having relatively \Red{much} smaller population) \\
Example: "The weather is warm because it is sunny".\\
(Elaboration: It is likely that the sunlight increases temperature. Yet there are many meteorology factors that affect temperature)
\end{tabular}
&
\begin{tabular}{p{.95\linewidth}}
\textbf{Specific law, regulation, convention, or written-down or by-the-book knowledge}\\
Reference: \citet{Lenat1995-vm} \\
Example: "Pistol is not prohibited in California".\\
(Elaboration: This statement expresses a regulation in the California State) \\
% Example: "A teacher teaches students in a class".\\
% (Elaboration: This statement expresses a convention of the job "teacher". It may or may not be the defining characteristics of the job, but it's an explicit widely-agreed derivative knowledge about the job "teacher")\\
\\
\textbf{Daily, encyclopedic, or scientific fact}\\
Reference: \citet{davis2023survey} \\
% Example: "The capital of the US is Washington D.C".\\
% (Elaboration: This statement expresses a encyclopedic fact (which is designated)) \\
Example: "The sun provides energy in 2 forms: heat and light".\\
(Elaboration: This statement expresses a scientific fact (which is proven))
\end{tabular}
\\

\bottomrule
\end{tabular}
}
\caption{Summary of assertions that concern either commonsense or referenced knowledge. Each collection (with \textbf{bold-faced} description) of instances is referred to previous works (Reference), and illustrated by an assertion (Example) and its elaboration (Elaboration).}
\label{tab:assumption}
% \vspace{-0.1in}
\end{table*}

In term of definition frameworks of concept, the assumption lies between and bridge the \textit{exemplar} and \textit{feature} frameworks. It not only regards the definition of concepts to instance level, but also summarizes instances into some representatives of the two concepts, which provide cues of the features by which we can discriminate the two concepts.

% Methodology:
% criteria proposal -> have reason to use self-proposed criteria, as many taxonomy-to-analysis papers also employ a blurry literature to propose the categories by themselves
% annotation process: the way that we use 3 existing views to annotate CSQA data, then parse it back to prototype view is reasonable?

\subsection{Features of Two Knowledge Types}
% https://poe.com/s/f9hmWbvISFWmEIj5fgjC

% acquisition:
%   sources: education, peer, self-redirected (must or can acquire through?)
%     active vs passive, public or private channel?
%   representation: implicit, explicit

% content and representation: social interaction, daily activities
%   science, encyclopedia, history, occasion

% scope + context + evaluation:
%   logical
%   law, convention, definition, linguistics, written down as (to some scale) empirically/clinically certified or proven
%   unproven mutual belief

% application -> from our point of view and previous work, not distinguishable from this aspect, as all knowledge is to be applied

% consult socialChemistry101 (okay, not useful for the discrimination)

To discriminate the two concepts, we dive deep into multiple aspects of knowledge, such as:
\begin{itemize}[leftmargin=*,label=$\bullet$,noitemsep,partopsep=0pt,topsep=0pt,parsep=0pt]
\item Acquisition: where to acquire the knowledge,
\item Content + Representation: which topics and objects does the knowledge regard,
\item Scope + Context + Evaluation: to what extent do people agree with or accept the knowledge (here, we do not consider ``knowing'', as it is subjective)
\end{itemize}

\noindent
For each of the aspects, we inherit the viewpoints from previous works and propose extra viewpoints as binary-value features to justify if one instance concerns commonsense or referenced knowledge. Depending on each feature, having the feature (i.e. having value 1) would make the instance inclined to commonsense or referenced knowledge (i.e. the knowledge type tendency of features). These features are summarized in Table \ref{tab:features}.

\begin{table*}[h!]
\small
\centering
\setlength{\tabcolsep}{4.5pt} % Default value: 6pt
\renewcommand{\arraystretch}{1} % Default value: 1
\resizebox{\linewidth}{!}{%
\begin{tabular}{p{0.12\linewidth} | p{0.55\linewidth} | p{0.1\linewidth} | p{0.03\linewidth}| p{0.2\linewidth}}

\toprule
Aspect $\downarrow$ & Feature & Notation & T. & Reference
\\

\midrule
\multirow{2}{*}{Acquisition} & Knowledge can be obtained through regular public channels (education, research, mass official media) & \texttt{a\_regular} & R & \citet{Lenat1995-vm}
\\
\cline{2-5} & Knowledge is often obtained through self-directed personal experience (before applying the knowledge or learning through public channels) & \texttt{a\_self} & CS & Complement of \texttt{a\_regular} 
\\

\midrule
\multirow{3}{*}{\tabincell{l}{Content + \\Representation}} & Knowledge regards to human daily interaction and activities, usage of human-created object, etc. & \texttt{cr\_social} & CS & \cite{sap2019atomic} and Complement of \texttt{cr\_stem}
\\
\cline{2-5} & Knowledge regards to STEM theory, nature, history, occasion, etc. & \texttt{cr\_stem} & R & \citet{davis2023survey}
\\
\cline{2-5} & Knowledge regards to a specific (named, identifiable) place, entity, object, method, number etc. & \texttt{cr\_spec} & R & \citet{Halpern1990-ng}
\\

\midrule
\multirow{3}{*}{\tabincell{l}{Scope + \\Context + \\Evaluation}} & Knowledge is logically/ empirically/ clinically certified or proven & \texttt{sce\_prov} & R & Derived from the word "referenced"
\\
\cline{2-5} & Knowledge is from law, convention, definition, scripts, linguistics & \texttt{sce\_conv} & R & \citet{Lenat1995-vm, davis2023survey}
\\
\cline{2-5} & Knowledge is of mutual subjective belief and observation & \texttt{sce\_heur} & CS & \citet{Liu2004-ag, davis2023survey}
\\
\bottomrule
\end{tabular}
}
\caption{The list of features we consider to distinguish commonsense from referenced knowledge. ``T.'' stands for tendency of knowledge type. In ``T.'' column, CS and R denotes commonsense and referenced knowledge.}
\label{tab:features}
\vspace{-0.1in}
\end{table*}

% These are the most representative among a long list of criteria that we hypothesize / propose, and attained via an iterative process of reflection and discussion on real textual data rather than a "magic" proposal. We provide more details about the full list of the criteria in Appendix ??. application -> from our point of view and previous work, not distinguishable from this aspect, as all knowledge is to be applied. Not mutually exclusive.

It can be observed that the features are not mutually disjoint in terms of semantics (even features in the same aspect). Also, this list may not be complete. However, as the first attempt to toward a comprehensive definition of commonsense, we only work on aforementioned aspects and features. 
The features are referenced or self-observed but all deemed subjective, thus, it is {unclear about the statistical significance of these features in determining which knowledge is commonsense}. Therefore, in the next subsection, we conduct a study on the CommonsenseQA and CommonsenseQA 2.0 datasets for examination. We explain our choice of datasets in Appendix \ref{sec:supplementary}.

% Thus we transform the description into attribute - value to clarify the definition. Where can we get the aspects? Which previous works derive an aspect? If it's proposed by us, what example motivate it?
% Aspect's transferability: use this to annotate other datasets (small set)? and compare with annotation on that dataset?? sound weird?

\subsection{Significance of Selected Features}
\label{sec:feature}
% To verify the self-derived theory

\paragraph{Expert Annotation}
We recruit three expert annotators who are postgraduate research students and having at least one year of research experience in the topic of commonsense. We randomly sample 100 instances from the development set of CommonsenseQA and CommonsenseQA 2.0, then ask three annotators to:
1) first, annotate the knowledge type of each instance (either commonsense or referenced),
2) then, after a week\footnote{According to \href{https://www.growthengineering.co.uk/what-is-the-forgetting-curve/}{the forgetting curve}, only 10\% of new things remains in humans' memory after a week.}, annotate the knowledge type tendency of each instance with respect to each feature.
The aggregated value of knowledge type and features are the majority among three annotations.  
% (use to educate/aid annotators, but not in this work, as seem not have any effect for expert annotation, prob. keep the annotation consistent)
% We note that the knowledge type of each instance in CommonsenseQA refers to the implicit knowledge that a person use to perform the task w.r.t. the instances. We discuss this choice of annotation objects as well as the choice of the dataset in Appendix \ref{sec:}.

% provide reason why we work on CSQA?
% Scope of research, on the dev split of CommonsenseQA, and judge the knowledge to answer the question give the question and the true answer (show how the setting different from other setting, e.g. only question, or assertion, or question + all answer. use case study to prove why we should take this setting)

\paragraph{Quality Control}
In term of quality control, we provide a full set of instructions including the motivation of this work, summary of existing definition frameworks of commonsense as distinguished from referenced knowledge (Tables ~\ref{tab:literature},\ref{tab:assumption}) with an extended set of examples. We carry out 3 hours of training and Q\&A session to familiarize annotators with the tasks. 
The fact that we let the annotators annotate the knowledge type before binary values of features is to mitigate any confounding factors other than the true relationship between them.
Based on heuristics, we argue that if annotators work on features before knowledge type, they would use the information of features to label the knowledge type, which incur trivial dependency.
Overall, for both datasets, the average Cohen's Kappa score w.r.t. the knowledge type and each feature are all greater than 0.4, which indicates at least moderate agreement. Among these data fields, knowledge type and \texttt{cr\_spec} feature of CommonsenseQA has the highest kappa 0.703 an 0.726, respectively. 

\paragraph{Feature Significance's Measurement}
Through regression models and decision tree models, we measure the statistical significance of features in determining which knowledge is commonsense. We set the knowledge type as the target, while features are attributes.
In term of regression models, we use linear models instead of logistic models, because 1) we can convert logistic models back to linear models through logarithm of the target and 2) linear models offers more numerically stable values of estimators and statistics. We apply the Backward Elimination Procedure (BEP) with metric AIC to select the best model, whose features are assumed by us to be the most significant in determining the knowledge type.
In term of the decision tree model, we fit the model (with the train/test ratio 8:2) in two settings: one includes all features, the other includes top 5 significant features (as features remained in the 5-feature model in the BEP).

The obtained result shows that for CommonsenseQA, \texttt{a\_self, sce\_convention, sce\_heuristic} are the most significant features (with p value at most 0.1), while the decision trees achieves 95\% prediction accuracy with the depth of 3 and the top 2 splits are based on \texttt{a\_self, sce\_heuristic}.
Likewise, for CommonsenseQA 2.0, \texttt{a\_regular, a\_self, sce\_heuristic} are the most significant feature (with p-values are 0, 0.117, 0 respectively), and the decision trees achieves 85\% the top 2 split nodes are based on \texttt{a\_regular, sce\_heuristic}.
We notice the difference in the significance of each feature in determining the knowledge type w.r.t. different datasets or data distribution in general, which is understandable.
However, features such as \texttt{a\_self} and \texttt{sce\_heuristic} are commonly significant, suggesting that we can use these features to quickly generalize to other data distribution and justify the knowledge type of new instances. 
In fact, value 1 of features \texttt{a\_self} and \texttt{sce\_heuristic} all indicate that the implicitness of a knowledge instance, as opposed to the explicitness of referenced knowledge.
We leave details about the regression models and decision tree models to Appendix \ref{sec:supplementary}.
% Dependency of each attribute w.r.t to others, coefficient significance! Leave the regression summary in Appendix
% Thus, we let the annotator work on the output first.

% Davis' work motivates us to do this research
% thus subsequently do corresponding experiments in step 1. study the perspectives of public (affect the annotation of commonsense instances (but in which situation that we need separate commonsense from referenced knowledge) -> ah, affect data curation, not the labeling step, make a more challenging step) and perspective of LLMs (just scale the annotation :v)
% how about testing our criteria on other datasets, e.g socialIQA?

\section{Analysis of Commonsense Datasets}

% \subsection{MTurk annotators' perspective. Need or not?}
% MTurk annotation, three setting as stated in Introduction?
% Since we don't test our criteria set, the perspectives we test is only for Turker.
% Compute the agreement between MTurk annotator and expert.  

% \subsection{Generalize to other datasets?}

\subsection{Fraction of Non-commonsense Instances in Commonsense Datasets}

By the annotation of knowledge type of 100 instances in subsection \ref{sec:feature}, we estimate the 95\% confidence interval of the proportion p of non-commonsense instances in CommonsenseQA and CommonsenseQA 2.0 datasets.
As the label is binary, we treat the knowledge type of an instance as a random variable of binomial distribution, with probability of {referenced} knowledge type (i.e. value 1) is p.\footnote{In fact, knowledge type can be in the form of typicality with continuous value, and it should follow a Gaussian distribution. However, it is difficult to quantify the typicality to continuous value.}
By our computation\footnote{We use this online statistics \href{https://www.socscistatistics.com/confidenceinterval/default2.aspx}{calculator}.}, the proportion p of non-commonsense instances in CommonsenseQA and CommonsenseQA 2.0 datasets are 0.27 ± 0.09 and 0.56 ± 0.1, which suggests the genuineness problem of the two datasets.

We also consider other famous commonsense datasets: WSC~\cite{levesque2012wsc} (one of the first commonsense benchmarks), HellaSwag~\cite{zellers2019hellaswag} (included in HELM Classic~\cite{liang2023holistic}),
% StrategyQA (one of two commonsense benchmarks in the suite of benchmarks to evaluate prompt engineering techniques, including Chain-of-Thought~\cite{})
and aNLI~\cite{bhagavatula2020abductive} (abductive reasoning).
By the nature of the data from WSC, every instance expresses a specific situation and the task as coreference resolution, which requires implicit (linguisitic) knowledge without any proven evidence.
Thus, these instances are deemed to be "Derivative of facts which is not written down and not always true" (Table \ref{tab:assumption}), which are assumed to be commonsense. That means the portion of non-commonsense knowledge instances in WSC is insignificant.
Likewise, aNLI concerns daily situation, thus the dataset is also likely genuine. We proof-check this heuristic by observing 50 random samples in each dataset, the result is as expected.
At the meantime, HellaSwag is constructed from two datasets ActivityNet~\cite{krishna2017densecaptioning} and WikiHow~\cite{koupaee2018wikihow}. While instances from ActivityNet is as situational as in aNLI, instances from WikiHow is not always commonsense but expert or specialized long-tail engineering knowledge. We evaluate 50 random samples with \texttt{sourceid} as WikiHow, and observe approximately half of those are non-commonsense.
% Last but not least, for StrategyQA, because each instance concerns a complex chain of reasoning, our theory does not apply to this dataset. Thus, we cannot make any argument. 

Overall, according to our consolidated definition of commonsense, there are so-called commonsense datasets with a large portion of non-commonsense instances, yet other commonsense datasets are of genuine commonsense. Nonetheless, our propositions are limited to knowledge type but not the task of the benchmarks. That means our theory has not yet rejected that these benchmarks do not concern commonsense reasoning.

% p = ; n = 100; math.sqrt(p*(1-p))
% csqa: 27 non-commonsense -> 
% csqa full 1221 instances: p = 0.2154, n =1221
% csqa2: 56 non-commonsense -> p = 0.56, sd = sqrt(p*(1-p)) = 0.4737, CI [0.5355, 0.7845].

% If sce_heuristic and a_self = 1, then very likely that the knowledge is commonsense. 
% WSC instances has such property. By 1-annotator inspection (like CREAK), WSC's instances mainly have value 1 on sce_heuristic, sce_linguistic (semantic but not definition), and a_self -> 85% accuracy with the decision tree means at least 85% of WSC is commonsense
% StrategyQA is too complicated to annotate. it concerns complex reasoning: https://allenai.org/data/strategyqa
% CREAK, Appendix B: not that "commonsense" as I thought. Prob. we can take a look at this dataset as well. But mixture of data type, hard to make a strong claim.
% HellaSWAG: take a look. Dataset is constructed based on ActivityNet and Wikihow, thus likely not proven, convention -> heuristic. Then it is not likely obtained via regular public (r u sure?), but self observation. Q? is there any instance with a_regular = 1, self_ob = 0, kinda expert knowledge. Likely no, as all situational. How about instances from Wikihow? Oh, at least 30% are specialized knowledge which is not common, thus for sure not commonsense knowledge

% Follow CREAK, u can do it by yourself as 1 annotator. Thus don't need official file 

\subsection{LLMs' Performance on Commonsense- and Referenced-Knowledge Instances}

For both CommonsenseQA and CommonsenseQA 2.0, we aim to compare the accuracy of LLMs on two subsets, one is on a subset consisting of commonsense-knowledge instances, the other is on a subset consisting of referenced-knowledge instances.
Following the prior analysis work~\cite{santy-etal-2023-nlpositionality}, we scale the annotation of knowledge type 
to 300 instances for CommonsenseQA and CommonsenseQA 2.0.

\begin{table}[h!]
\small
\centering
\setlength{\tabcolsep}{4.5pt} % Default value: 6pt
\renewcommand{\arraystretch}{1.1} % Default value: 1
\begin{tabular}{c|cccc}
\toprule
Model $\to$ & Gemini & ChatGPT & LLaMa2 & Mixtral \\
\midrule
\multicolumn{5}{l}{\textbf{CommonsenseQA}} \\
\midrule
Commonsense & 75.53 & 75.10 & 61.37 & 72.10 \\
Referenced & 80.59 & 76.11 & 68.65 & 79.10 \\
\midrule
\multicolumn{5}{l}{\textbf{CommonsenseQA 2.0}} \\
\midrule
Commonsense & 70.90 & 66.36 & 43.63 & 60.90 \\
Referenced & 74.73 & 64.21 & 47.36 & 67.89 \\
\bottomrule
 
\end{tabular}
\caption{LLMs' performance with Accuracy metric.}
\label{tab:llm-performance}
\vspace{-0.1in}
\end{table}

We employ four LLMs, which are Gemini-Pro, ChatGPT (Jun 2023 version), LLaMa2-7B-chat, and Mixtral-8x7B, 
% For Gemini, ChatGPT, and Mixtral, we call the models via APIs. For LLaMa2, we deploy it on own server.
as they are available, stable, and four of the most capable models at the time we were conducting our experiments. We set temperature T = 0 and use zero-shot prompt for all generation. The prompt instruction for each task is adapted from HELM.
The result is shown in Table~\ref{tab:llm-performance}. We notice a significant performance gap (varying from 4 points  to 7 points accuracy) between the performance of LLMs, except ChatGPT, on the two subsets of both datasets.
It suggests that tasks involving commonsense knowledge (or in a possibly overclaimed term, commonsense reasoning tasks) are more challenging than tasks involving referenced knowledge, which concerns memory retrieval.
In term of ChatGPT, we argue that because OpenAI collects conversation data to further train the models and there are a lot of commonsense benchmarks are used to evaluate ChatGPT, ChatGPT is well-trained with verbalized commonsense knowledge.

% Furthermore, as an attempt to demystify the performance gap, we examine the perplexity of instances from two subsets w.r.t. LLaMa2-7B-chat. We compute the language modeling loss\footnote{via Huggingface Transformers implementation.} of assertions that back the instances, and obtain the average loss for commonsense and referenced-knowledge subsets of CommonsenseQA/2.0 are 4.38/4.30 and 4.30/4.15 respectively. That supports our argument in subsection \ref{sec:review} that referenced knowledge is inherently reported more frequently than commonsense knowledge, and because of that, LLMs perform better in referenced-knowledge instances.

% No, as the question is not the knowledge, thus the question may be weird thus having high perplexity but it just need knowledge retrieval.
% But maybe in this work, we don't try to explain yet. Just pose the problem! 
% Link it to the Head-to-Tail Knowledge on LLMs.

% Use this test https://www.socscistatistics.com/tests/studentttest/default.aspx
% May be this calculator is more suitable https://select-statistics.co.uk/calculators/two-sample-t-test-calculator/
% Hmm, not our data will not satisfy the assumptions. But still need to use!
% 100 instances -> not significant 95\%
% 200 instances -> .. hmm, even worse!
% Seem that I need to find other type of statistical test.

\section{Conclusion}

In this work, we demystify claims regarding the genuineness of commonsense datasets. We survey and consolidate existing definitions of commonsense knowledge through the three frameworks for defining concepts.
We then use the consolidated definition to show that there exists a large portion of non-commonsense knowledge in CommonsenseQA and CommonsenseQA 2.0. There is also a large performance gap on two subsets of commonsense and referenced knowledge in the two datasets, where LLMs perform worse on commonsense-knowledge instances.
Although we do not long for perfect commonsense datasets, our work aims to raise the awareness of the genuineness problem of commonsense datasets. In general to the NLP community, we call for theoretical works on subfields of NLP research which deal with unclear concepts. That would facilitate better understanding of the underlining problems for the NLP community.

% "A full understanding of any text then, requires a surprising amount of commonsense, .." (ConceptNet)
% Discuss the role of commonsense and referenced knowledge in real life, common -> book learning, commonsense -> observation. In many unexpected case, commonsense would lead to better result. Kinda argument
% No benchmark, no taxonomy is perfect. And we don't need it

\newpage
\section*{Limitation}

This paper works on the definition of commonsense as distinguished from referenced knowledge - the cover of common, encyclopedic, and expert knowledge, and based on that, it demystifies claims concerning commonsense.
Due to limited human resource, only a few datasets are empirically studied and the obtained results are unavoidably subjective in a certain level. Further study with a larger scale is expected to examine the generalizability of insights drawed from this work.
Also, this work discusses the blurry concept of commonsense from perspectives of annotators and researchers working on commonsense, which possibly lead to the genuineness problem of commonsense datasets, however, there is no empirical study as attempt to clarify the cause. Experiments with researchers from various research groups and crowdsourced annotators (e.g. in Amazon Mechanical Turk) are preferred to make the arguments in this work more convincing.

\section*{Ethical Statments}

This work provides a (more) comprehensive literature of the concept ``commonsense'' by examining and experimenting with many commonsense datasets and benchmarks. Thus, this work shares the same ethical issues as these previous works. By our inspection, all sampled data instances do not contain any private information about any specific entities (e.g., a person or company). We carried out human expert annotation, where annotators are fairly paid according to the minimum wage requirement of the local government.

In another aspect, the study of LLMs' performance on data subsets of different knowledge types involves the use of Gemini (gemini-pro), ChatGPT (gpt-3.5-turbo-0613), LLaMa2 (llama-2-7b-chat-hf), and Mixtral (mixtral-8x7b-instruct). Except LLaMa2 which is deployed on local server, other three LLMs are called via APIs provided by GoogleAI, OpenAI, and Fireworks.AI respectively. Thus, the same risks from LLMs research are applicable to this work~\cite{Bender2021-xz}.

\bibliography{custom}
\bibliographystyle{acl_natbib}

\appendix
\newpage

\section{Discussion on Knowledge Types}
\label{sec:discussion}

\subsection{General}
%% Discuss in previous work or literature review already

% Can we unify commonsense and referenced knowledge? In term of logical thinking, any inference can be broken into axioms/facts/common believes. Since facts are what we discovered from nature at this moment from our perspective, axioms are what we defined (linguistic, laws, ...), while common believes are what we shared and followed together but not be always true by nature or conventionalized.

% Basic units of reasoning/logic.

% This work is not meaningful if people only treat AI models as tool and stop debating about capability of models philosophically.

% Learn as fact in an event -> Generalize as commonsense in general situation

Although we aim to discriminate commonsense from referenced knowledge, we are not strict in the discrimination (and almost impossible to do it perfectly). It is undeniable that the knowledge type is varying between different people and even different timestamps of a person. We name it the circulation of knowledge, or the journey from unknown to known and popular of a knowledge. A person may learned a popular knowledge from official mass media directly, or learned a knowledge from observation (thus treat it as commonsense), then the knowledge may be conventionalized to be a referenced knowledge. A referenced knowledge that a person does not know may be commonsense for them.

About our decision to group common, expert, and encyclopedic knowledge into the same categories; we observe the dynamics of these knowledge types through time. An expert/encyclopedic knowledge can be popularized, making it more "common" to public. Also, a common knowledge for a person may be expert knowledge for others, as different people have different expertise. In fact, by our annotation on CommonsenseQA, the proportion of encyclopedic/expert knowledge is approximately 5\%. Therefore, we merge them to a group, named \textbf{referenced} knowledge, as the knowledge is true by nature or conventionalized and it need a \textbf{reference} for its validity.

Furthermore, we want to relate commonsense and referenced knowledge in our work to other concepts. About the data frequency or distribution, we aforementioned, commonsense tends to be more ``long-tail'' than referenced knowledge. In term of reasoning, commonsense likely exists and be necessary for with abductive reasoning, while referenced associate with logical reasoning. Likewise, commonsense knowledge represents a probabilistic world, and referenced knowledge bases a deterministic world.

% \subsection{Case Studies on Subtle Difference of Commonsense and Referenced Knowledge}

% We discuss several cases that consist of relevant assertions but of different knowledge type (according to our consolidated theory, at least) to show the subtle difference of commonsense and referenced knowledge.

% \noindent
% \textbf{Case 1.} Interaction between type of questions.
% \begin{itemize}[leftmargin=*,label=$\bullet$,noitemsep,partopsep=0pt,topsep=0pt,parsep=0pt]
% \item If you spend a long time running, you likely feel tired (commonsense subjective feeling)
% \item Increased heart rate is an unavoidable physiological consequence of running (referenced biology knowledge)
% \end{itemize}

% \textbf{Case 2.} Interaction of fact and `derivative' of fact.
% \begin{itemize}[leftmargin=*,label=$\bullet$,noitemsep,partopsep=0pt,topsep=0pt,parsep=0pt]
% 1. Joe plays a percussion instrument in something. What might be play in? orchestra: a situation, thus forward (commonsense)
% 2. Where would you hear a violin along side many string and wind instruments? orchestra: backward, as asked about a place that has this property (common)
% 3. Where would you hear a trumpet along with other instruments made from the same material? brass band (common)
% \end{itemize}

\section{Further Discussion and Supplementary Materials}
\label{sec:supplementary}

\subsection{Choices of Datasets and Annotation Objective}

There are several prevalent (textual) commonsense datasets such as ATOMIC, SocialIQA, SocialChemistry101, etc., however, we choose CommonsenseQA and CommonsenseQA 2.0 because from observation from a previous work and ourselves, they contain substantial amount of non-commonsense data. Also, the tasks of the two datasets are different, thus, we can study how different are the contribution of/correlation between tendencies in criteria and the knowledge type.

In term of the annotation objective, as an assertion may carry several piece of knowledge, we determine the knowledge type of an assertion based on the top non-grammatical division of the corresponding dependency tree, or the division which supports the answer for the task.
In case a instance is not accurate\footnote{The accuracy of CommonsenseQA dataset is not close to 100\%, admitted by the authors} or widely known, we assume it is correct and widely known as the standard data properties for annotation.

\subsection{Evaluation of Features}

We show the agreement of annotations of features' value in Table \ref{tab:feature-annotation-agreement}.

\begin{table}[h!]
\small
\centering
\setlength{\tabcolsep}{4.5pt} % Default value: 6pt
\renewcommand{\arraystretch}{1.1} % Default value: 1
\begin{tabular}{l|ccc|c}
\toprule
K. Type/ Features $\downarrow$ & A01 & A12 & A20 & Avg \\
\midrule
\multicolumn{5}{l}{\textbf{CommonsenseQA}} \\
\midrule
Knowledge Type & 0.6896 & 0.595 & 0.8246 & 0.7030 \\
\texttt{a\_regular} & 0.6664 & 0.5949 & 0.7045 & 0.6552 \\
\texttt{a\_self} & 0.3316 & 0.5949 & 0.6286 & 0.5183 \\
\texttt{cr\_social} & 0.5821 & 0.6712 & 0.6271 & 0.6267 \\
\texttt{cr\_stem} & 0.4565 & 0.4052 & 0.6191 & 0.4936 \\
\texttt{cr\_spec} & 0.8077 & 0.734 & 0.6355 & 0.7257 \\
\texttt{sce\_prov} & 0.4492 & 0.6009 & 0.6491 & 0.5664 \\
\texttt{sce\_conv} & 0.4334 & 0.2982 & 0.6721 & 0.4679 \\
\texttt{sce\_heur} & 0.3735 & 0.4699 & 0.7333 & 0.5255 \\
\midrule
\multicolumn{5}{l}{\textbf{CommonsenseQA 2.0}} \\
\midrule
Knowledge Type & 0.376 & 0.5881 & 0.4367 & 0.4669 \\
\texttt{a\_regular} & 0.511 & 0.4538 & 0.5264 & 0.4970 \\
\texttt{a\_self} & 0.6502 & 0.4689 & 0.375 & 0.4980 \\
\texttt{cr\_social} & 0.52 & 0.54 & 0.502 & 0.5206 \\
\texttt{cr\_stem} & 0.3681 & 0.4694 & 0.3844 & 0.4073 \\
\texttt{cr\_spec} & 0.6523 & 0.6689 & 0.559 & 0.6267 \\
\texttt{sce\_prov} & 0.428 & 0.5407 & 0.3409 & 0.4365 \\
\texttt{sce\_conv} & 0.5403 & 0.4749 & 0.5588 & 0.5246 \\
\texttt{sce\_heur} & 0.5339 & 0.3471 & 0.4813 & 0.4541 \\
\bottomrule
 
\end{tabular}
\caption{Agreement of annotations of features' value and knowledge type. A\{i\}\{i+1 mod 3\} denotes the Cohen's Kappa of annotations of (i+1)- and (i+2)-th annotators. Avg. denotes the average Cohen's Kappa of three annotators.}
\label{tab:feature-annotation-agreement}
\vspace{-0.1in}
\end{table}

\begin{figure}[h!]
    \centering
    \includegraphics[width=1\linewidth]{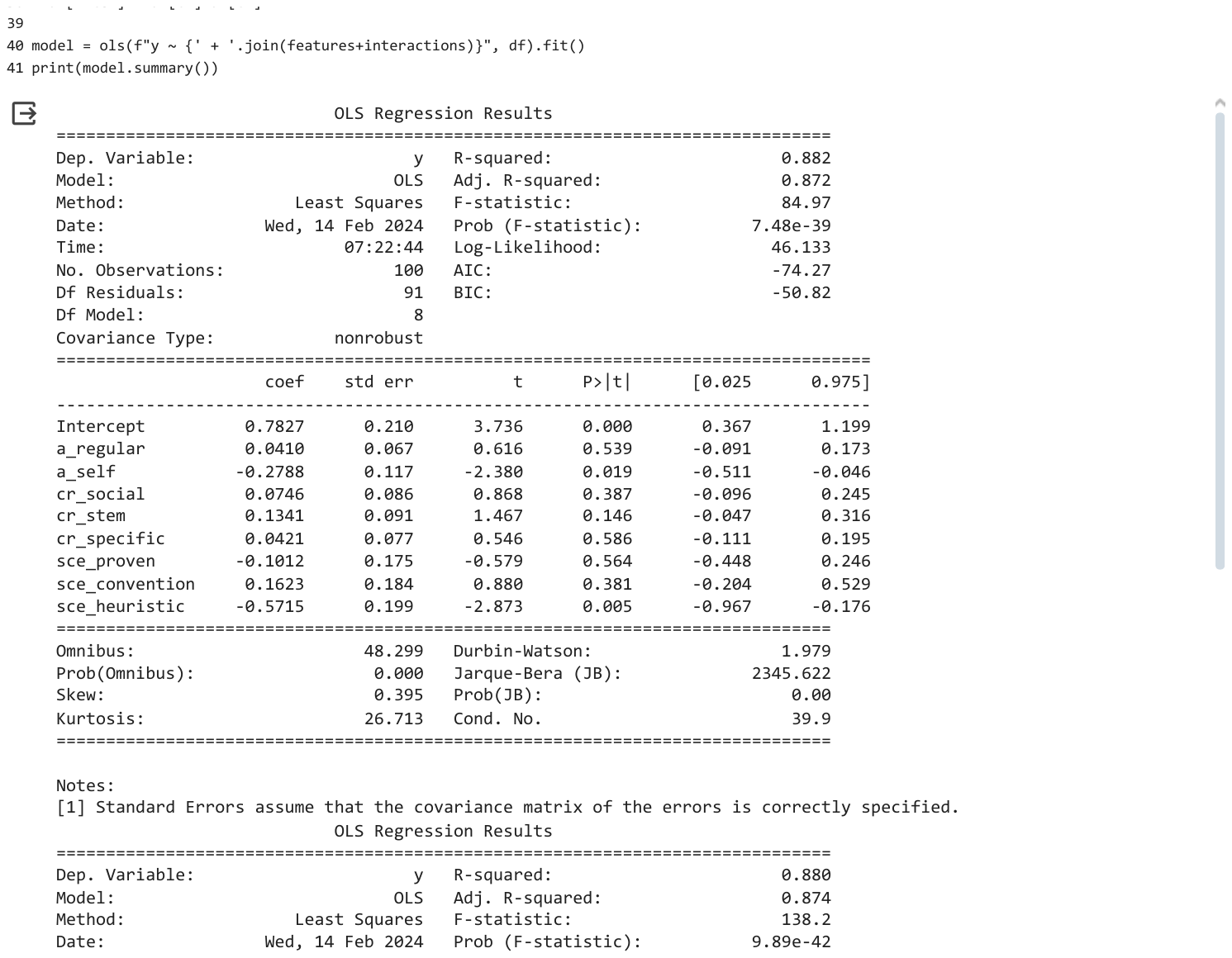}
    \includegraphics[width=1\linewidth]{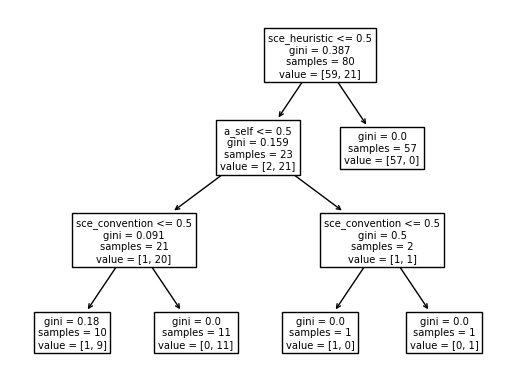}
    \includegraphics[width=1\linewidth]{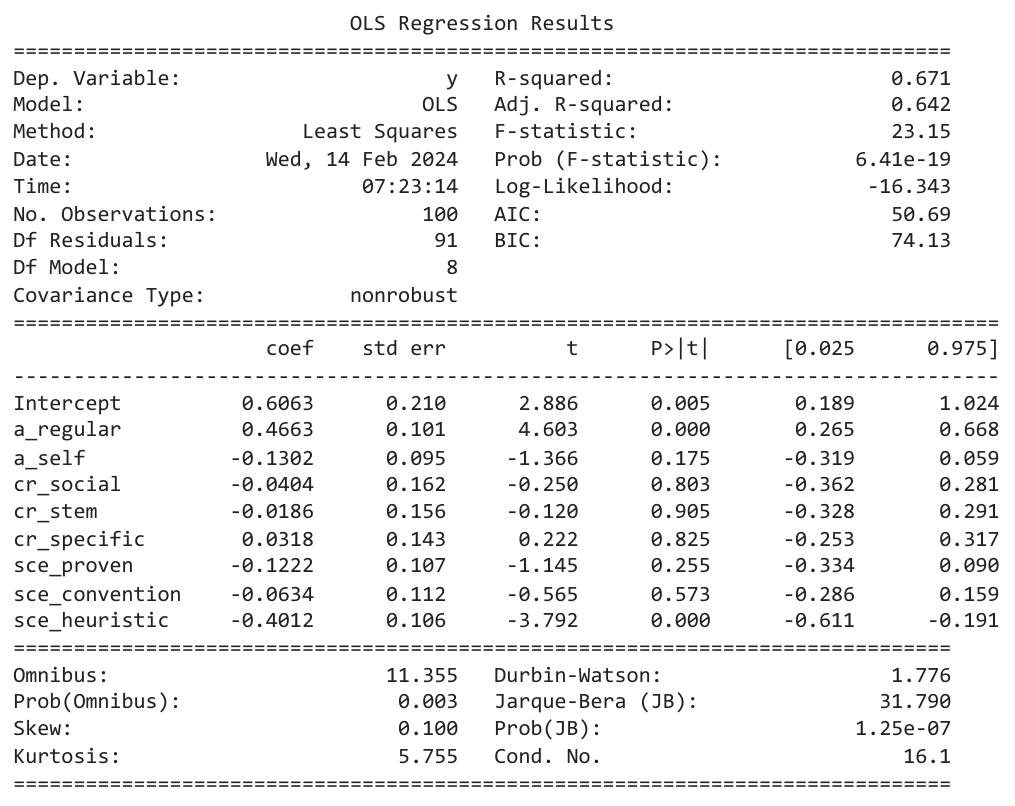}
    \includegraphics[width=1\linewidth]{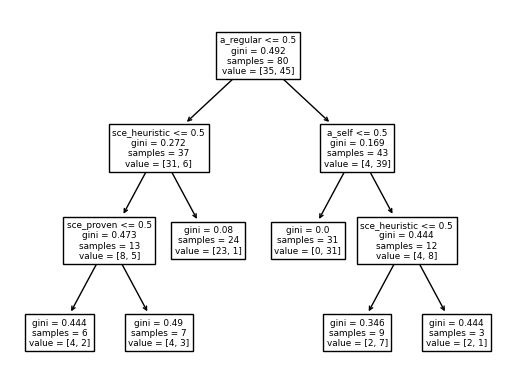}
    % \vspace{-0.2in}
    \caption{Details of the full-feature regression model as well as decision tree of CommonsenseQA and CommonsenseQA 2.0}
    \label{fig:stat}
    \vspace{-0.1in}
\end{figure}

\noindent
Next, we show the statistics of features' significance via the full-feature regression model as well as decision tree of CommonsenseQA and CommonsenseQA 2.0 respectively, in Figure~\ref{fig:stat}.

\subsection{Better Guideline for Annotation}

Based on the consolidated definition commonsense, one may rely on the proximity of new instance to the representative cases in Table \ref{tab:assumption} to decide the knowledge type. Meanwhile, for uncertain cases, one can work with the feature lists in Table \ref{tab:features}. Note that the significance of features may vary according to data distribution, thus a pilot study of the significance of features is preferred for better judgment. Nonetheless, 1) whether we can obtain the knowledge by our own experience/observation and 2) whether the knowledge is only mutual belief are the most significant features to identify commonsense from referenced knowledge.

\subsection{Portion of Non-Commonsense Instance}

We are aware of the p-hacking problem with sampled data, thus we extend the annotation of knowledge type on CommonsenseQA to its whole development set which consists of 1221 instances and recompute the confidence interval to prove the representativeness of our sampled data.
We get 0.2153 ± 0.0231, which is expectedly not varying much in term of lower bound in comparison to previously computed confidence interval.
Also, as some people may argue the subjectivity in annotation of basic knowledge which lies between of commonsense and referenced knowledge, we analyze the portion of encyclopedic and expert knowledge which are certainly non-commonsense. Considering the whole development set of CommonsenseQA, there are approximately 5\% of the instances which concern encyclopedic and expert knowledge.

\subsection{Evaluation of LLMs on Two Subsets}

We want to treat the accuracy as the mean of random variables from two populations (i.e. two subsets), whose random variables are indicator functions with each representing if an LLM's answer to a task instance is correct, and conduct the Two-sample t-test.
However, the distribution of our data is binomial, which does not satisfy the assumption of data normality of the test. Therefore, we only compare the accuracy in a straight forward manner.

\end{document}